\documentclass{article} % For LaTeX2e
\usepackage{iclr2020_conference,times}

\usepackage{amsmath,amsfonts,bm}

\def\eqref#1{equation~\ref{#1}}
\def\1{\bm{1}}

\DeclareMathAlphabet{\mathsfit}{\encodingdefault}{\sfdefault}{m}{sl}
\SetMathAlphabet{\mathsfit}{bold}{\encodingdefault}{\sfdefault}{bx}{n}

\newcommand{\R}{\mathbb{R}}

\usepackage{hyperref}
\usepackage{url}
\usepackage{textcomp}
\usepackage{amsmath}
\usepackage{flexisym}
\usepackage{caption}

\usepackage{subcaption}
\usepackage{graphicx}   %% no demo in your file
\usepackage{algorithm}
\usepackage[noend]{algpseudocode}
\usepackage{pgfplots, tikz}
\pgfplotsset{yticklabel style={text width=3em,align=right}}
\title{Domain Knowledge Integration by Gradient Matching for Sample-Efficient Reinforcement Learning}
\author{Parth Chadha\\
\texttt{parth29@gmail.com}
}
\iclrfinalcopy % Uncomment for camera-ready version, but NOT for submission.
\begin{document}

\maketitle

\begin{abstract}
Model-free deep reinforcement learning (RL) agents can learn an effective policy directly from repeated interactions with a black-box environment. However, in practice the algorithms often require large amounts of training experience to learn and generalize well. In addition, classic model-free learning ignores the domain information contained in the state transition tuples. Model-based RL, on the other hand, attempts to learn a model of the environment from experience and is substantially more sample efficient, but suffers from significantly large asymptotic bias owing to the imperfect dynamics model. In this paper, we propose a gradient matching algorithm to improve sample efficiency by utilizing target slope information from the dynamics predictor to aid the model-free learner. We demonstrate this by presenting a technique for matching the gradient information from the model-based learner with the model-free component in an abstract low-dimensional space and validate the proposed technique through experimental results that demonstrate the efficacy of this approach. 
\end{abstract}

\section{Introduction} 
% model-free methods are powerful
% but model-based information helpful in planning etc.
Reinforcement Learning (RL) methods aim to solve the fundamental problems involving \textit{learning} and \textit{planning}, where the \textit{learning} problem aims to improve the policy via repeated interactions with the environment, and \textit{planning} improves the policy without interaction with the environment. These problems further divide into two sub-domains of RL, \textit{model-free} and \textit{model-based} methods. Model-free RL using expressive deep learning techniques has recently achieved great success in several game playing tasks (\cite{mnih2016asynchronous}; \cite{mnih2013playing}; \cite{silver2016mastering}; \cite{sutton2000policy}), however, it often requires large amounts of interaction with the environment and doesn't generalize across tasks with similar environmental dynamics. On the other hand, model-based RL (\cite{sutton1991dyna}) attempts to learn a model of an environment from experience gained during training and uses this model to plan ahead. 

It has been shown recently that a combined approach incorporating both the model-based and model-free components can be used to solve problems which are unsolvable if either of the component is used in isolation (\cite{franccois2019combined}; \cite{silver2017predictron}; \cite{oh2017value}). These architectures train both components end-to-end, often performing planning in a low-dimensional abstract space. While prior work has shown the increased capability of the agent in an online setting by using both components, it is still difficult for agents to learn from limited data and generalize across a variety of tasks. Recent work on Combined Reinforcement Learning via Abstract Representations (CRAR) (\cite{franccois2019combined}) has addressed this issue by using the agent in a meta-learning setting with limited off-policy data. 

In this work, we propose an algorithm for Gradient-Matching in the Abstract Space (GMAS) to incorporate domain knowledge acquired by the model-based learner as a means to facilitate faster learning and better generalization in unseen environments in which the underlying dynamics are similar to the training environment dynamics. Specifically, by providing an additional training signal using a loss based on the objective function gradient mismatch with respect to the abstract representation space between the model based planner and the model free learner, we demonstrate faster learning and improved test performance in a meta learning setting composed of environments drawn from a distribution of labyrinth maze tasks. In addition, through a set of ablation experiments, we demonstrate experimentally the effect of planning depth for GMAS on sample efficiency and discuss possible extensions of the work to mitigate compounding errors introduced by model based planners.
% Incorporating information model-based learning information with model-free learner helps in generalization performance as long as the underlying task dynamics remain consistent. 

% second para : latent space planning; sample efficiency problem explain; explain ebnn technique
% third para : 
\section{Background}
%In this section we provide background on combined model-free and model-based learning for RL. 
The reinforcement learning setting involves an agent learning to act in a Markov Decision Process (MDP) to maximize the expected discounted return $R_{t}$ where $R_{t} = \sum_{t=0}^{\infty} \gamma^{t} r_{t} $ by optimizing its policy $\pi(s)$ that maps states to actions where state $s\in \mathcal{S}$ and action $a\in \mathcal{A}$. In model-free RL, Q learning aims to find the optimal action value function (Q function, where $Q : \mathcal{S} \times \mathcal{A} \rightarrow R $)  which gives the best value possible from any policy, i.e., $Q^{*}(s,a) = \mathop{\mathbb{E}}_{s'}[r(s,a) + \gamma \max_{a'} Q^{*}(s',a')]$, where $r$ is the reward function, $\gamma$ the discount factor and the expectation depends on the transition function $T(s'|s,a)$. Model based RL methods aim to directly learn the transition function from the observed roll-outs during training. These methods also aim to predict the reward at each time step as well as whether a state is terminal or not. 
% In this section we provide background on planning, value iteration, CNNs, and policy representations for RL and IL. In the sequel, we shall show that CNNs can implement a particular form of planning computation similar to the value iteration algorithm, which can then be used as a policy for RL or IL.Va
\subsection{Combined Reinforcement Learning via Abstract Representations}
\label{crar}
\cite{franccois2019combined} introduce the CRAR agent that combines model-based and model-free components using a loss function that encourages both components to share a common underlying abstract representation space, demonstrating improved generalization while also providing computational benefits for efficient planning. The model uses an encoder $\textit{e}$ which maps the raw observation $s$ from the environment to an abstract space $x \in \mathcal{X}$ such that both the model-free and model-based components operate on $x$. While the model-free learner uses double-Q learning (\cite{van2016deep}) to learn an efficient policy, the model-based component updates its models for the reward $\rho(s,a; \theta_{\rho})$, discount factor $g(s,a;\theta_{g})$ and transition model $\tau(s,a;\theta_{\tau}))$ according to the loss functions described below using the tuple $(s,a, r, \gamma, s\textprime)$ from the replay buffer. The authors use the discount factor $\gamma$ to identify terminal states for planning (details in original work).
\begin{equation}
\begin{split}
    L_{\rho}(\theta_{e}, \theta_{\rho}) = \left|r-\rho(e(s;\theta_{e}),a; \theta_{\rho}) \right|,
    \\
    L_{g}(\theta_{e}, \theta_{g}) = \left|\gamma-g(e(s;\theta_{e}),a; \theta_{g}) \right|, 
    \\
    L_{\tau}(\theta_{e}, \theta_{\tau}) = \left|e(s;\theta_{e}) + \tau(e(s;\theta_{e}), a; \theta_{\tau})  - e(s\textprime;\theta_{e}) \right|   
\end{split}
\end{equation}
The gradients from the losses above are back-propagated through both the common encoder and the individual components. To prevent contraction of the abstract representation, an auxiliary loss is used for entropy maximization as shown in Equation \ref{em} and another loss constrains the abstracts representation within an $L_{\infty}$ ball of radius 1 to prevent very large values. 
\begin{equation}
\label{em}
    L_{d1} (\theta_{e}) = \exp(-C_{d} \lVert e(s1;\theta_{e}) - e(s2;\theta_{e}) \rVert)
\end{equation}
\begin{equation}
\label{linf}
    L_{d2} (\theta_{e}) = \max (\lVert e(s1;\theta_{e})  - 1 \rVert_{\infty}, 0)
\end{equation}
At evaluation time, the agent utilizes both model-free and  model-based components to estimate the optimal action at each time step. The agent performs an expansion step up to planning depth $D$. At each step it evaluates $b_{d} \in \mathcal{A}$ actions, where $b$ is a hyper-parameter chosen based on the environment. A backup step utilizes the trajectories with the best return to estimate the Q-value at depth d ($Q^{d})$. Finally, the agent chooses an action using the average of the Q values starting from the model-free estimate ($d=0$) up to the model-based estimate till planning depth $D$ : $Q^{d}_{plan}(x,a) = \frac{\sum_{d=0}^{D} Q^{d}(x,a)}{D}$.
% \subsection{Explanation-Based Neural Network Learning}
%\pagebreak
\section{Methods}
Here we explain how the model-based planner can be used to better guide the model-free learner towards learning an optimal estimate of the Q-value function. As described in \ref{crar}, the CRAR agent uses model-based components for planning and estimating the Q-value function. The depth-d estimated expected return can be defined as :
\begin{equation}
\label{plan}
   % \[
    \hat{Q}^{d}(x_{t},a)= 
\begin{cases}
    \rho(x,a) + g(x,a) \times \max_{a' \in \mathcal{A^{*}}}Q^{d-1}(x_{t+1},a') ,& \text{if } d > 0\\
    Q(x_{t},a;\theta_{k}),              & \text{if } d = 0
\end{cases}
%\]
\end{equation}

\subsection{Gradient Matching Algorithm}
% describe why gradient information from aux task can help
% explain what model-based is doing and planning eq's
% show the calculation of gradient, loss fun, l2 vs cosine, use of target function
The goal of GMAS is to maximize the correlation between the slopes (partial derivative of the objective function with respect to input) of the model being trained and an auxiliary trained model operating under similar environmental dynamics. This approach was first described in \cite{mitchell1993explanation}. For example, consider a dataset $D$ with input features $x^{(i)} \in \R^{n}$ and output $y \in [0,1]$. While training a model $M$ parameterized by $\theta$ on $(x_{n}, y_{n})$, we provide an additional target label $\nabla_{x} y_{n}$ which is estimated using a separate neural network trained on an auxiliary task drawn from the same distribution as the main task. This additional label information for each data point helps in improving the sample efficiency of the learner. 

For the case of an RL algorithm composed of both model-free and model-based components, the model-based learner acts as a domain knowledge expert and provides an approximation of the underlying environmental dynamics thereby aiding generalization by utilizing task-independent knowledge to bias the model-free learner.
Specifically, the slope of Q-value estimated at depth $d$ as described in Equation \ref{plan} is
\begin{equation}
\label{der}
%    \[
    \frac{\partial \hat{Q}^{d}(x_{t},a)}{\partial x}= 
\begin{cases}
    \frac{\partial \rho(x_{t},a)}{\partial x_{t}} + \frac{\partial g(x_{t},a)}{\partial x_{t}} \times \max_{a' \in \mathcal{A^{*}}}Q^{d-1}(x_{t+1},a') + \\ g(x_{t},a) \times \frac{\partial Q^{d-1} (x_{t+1},a'')}{\partial x_{t}} \times \frac{\partial \tau(x_{t},a'')}{\partial x_{t}} & \text{if } d > 0\\
    \\
    \frac{\partial Q(x_{t},a;\theta_{k})}{\partial x_{t}}   & \text{if } d = 0
\end{cases}
%\]
\end{equation}
and the gradient of the complete model-based planner is $\frac{\partial \hat{Q}^{d}_{plan}(x_{t},a)}{\partial x} = \sum_{d=0}^{D} \frac{\partial \hat{Q}^{d}_{plan}(x_{t},a) \mathbin{/} \partial x}{D}$. Equation \ref{der} requires the best action $a''$ for each planning depth $d$ and this is found by the backup stage of model-based planner.
GMAS appends  $\frac{\partial \hat{Q}^{d}_{plan}(x_{t},a)}{\partial x}$
to the training data tuple of the double-DQN agent thereby helping in a) sample efficiency by providing more information per data point and b) better generalization due to goal-agnostic dynamics information. In CRAR, the loss function of the model-free component is
\begin{equation}
    \mathcal{L}_{modelfree} (\theta_{e}, \theta_{Q}) = (Q(e(s;\theta_{e}),a;\theta_{Q}) - Y)^2
\end{equation}
where, $Y$ is the target value under the double-DQN algorithm (for more details refer to \cite{franccois2019combined}).
With GMAS, this becomes
\begin{equation}
    \mathcal{L}_{modelfree} (\theta_{e}, \theta_{Q}) = (Q(e(s;\theta_{e}),a;\theta_{Q}) - Y_{k})^2 + \alpha \times dist(\frac{\partial \hat{Q}^{d}_{plan}(x_{t},a)}{\partial x}, \frac{\partial \hat{Q}^{d}_{modelfree}(x_{t},a)}{\partial x})
\end{equation}
where $\alpha$ is a scaling factor and $dist$ is a generic distance function and in this work we explore $L2$ norm and cosine distance ($1-\frac{\sum_{i=1}^{n}x_{i}y_{i}}{\sqrt{\sum_{i=1}^{n}x_{i}^2} \sqrt{\sum_{i=1}^{n}x_{i}^2}}$) as shown in section \ref{exp}.

%MENTION THE USE OF OLD Q FUNCTION INSTEAD OF NEW ONE TO PREVENT MOVING TARGET PROBLEM.
\subsubsection{Gradient with respect to the abstract space}
A key aspect of GMAS is that all partial derivatives are calculated with respect to the abstract representation $x$ and not the input state $s$. This choice is crucial for two reasons. First, $s$ is high dimensional whereas $x$ is its compact representation capturing only salient information. The sparse gradient with respect to $s$ limits the applicability of the distance metric. Second, since the abstract space is shared between the model free and model based learners (and is used for planning ahead), it provides a  constrained generalized feature representation, making it a better choice for GMAS.
\subsubsection{Combating the moving target problem and compounding error}
\label{compound}
The estimation of gradient information is dependent on the Q value function (Equation \ref{der}). This causes a moving target problem since the parameters of the Q function are updated with each gradient step. To tackle this, we use the frozen target Q-function ($\theta_{Q}^{-}$) for the estimate of $\frac{\partial \hat{Q}^{d}_{plan}(x_{t},a)}{\partial x}$, similar to its usage in \cite{mnih2013playing}.

%\subsubsection{Usefulness of the Gradient information}
%higher order gradients?!
The dependence on the model-based planner for gradient estimates leads to a compounding-error problem as discussed in \cite{jiang2015dependence}, \cite{talvitie2017self} and \cite{wang2019benchmarking}. The error of the gradient estimate increases with the planning depth in case of an imperfect dynamics model. We discuss below possible methods to counter the compounding error problem.
\begin{itemize}
    \item Calculating the deviation of the model-based component with respect to the ground-truth and factoring the deviation in the slope calculation (as discussed in \cite{franccois2019combined}). This method has drawback in the case of off-policy meta-learning task discussed in this work due to the dependence on the latest policy to calculate the deviation. 
    \item Calculating planning depth on-the-fly using Adaptive Model-based Value Expansion \cite{xiao2019learning}.
    \item Using a fixed discount factor $\gamma'$ to discount the estimate of gradient from far-away depths.
\end{itemize}
% n order to make a prediction about the future, dynamicsmodels are unrolled step by step which leads to “compound-ing errors” (Talvitie, 2014; Bengio et al., 2015; Lamb et al.,2016): an error in modeling the environment at timetaffectsthe predicted observations at all subsequent time-steps
\section{Experiments}
\label{exp}
We evaluate our algorithm on tasks sampled from a distribution of maze environments, where a random maze is generated each time the environment is reset. The agent is restricted to learn with limited off-policy data collected with a random policy. The agent receives a reward of $+1$ when it reaches the key and $-0.1$ for all other transitions. During evaluation of the learned policy, an episode is considered over after 50 steps or if the agent receives all the rewards. The same environment setup was used to evaluate the CRAR agent in \cite{franccois2019combined} and hence we compare the improvement from GMAS with the original baseline for cases where the CRAR agent performs poorly. 

% \subsection{Effect of planning depth}

%\subsection{Combined Reinforcement Learning Experiments}

\begin{figure*}[h!]
    \begin{subfigure}{0.24\textwidth}
    \centering
    \includegraphics[width=\textwidth, height=1.2in]{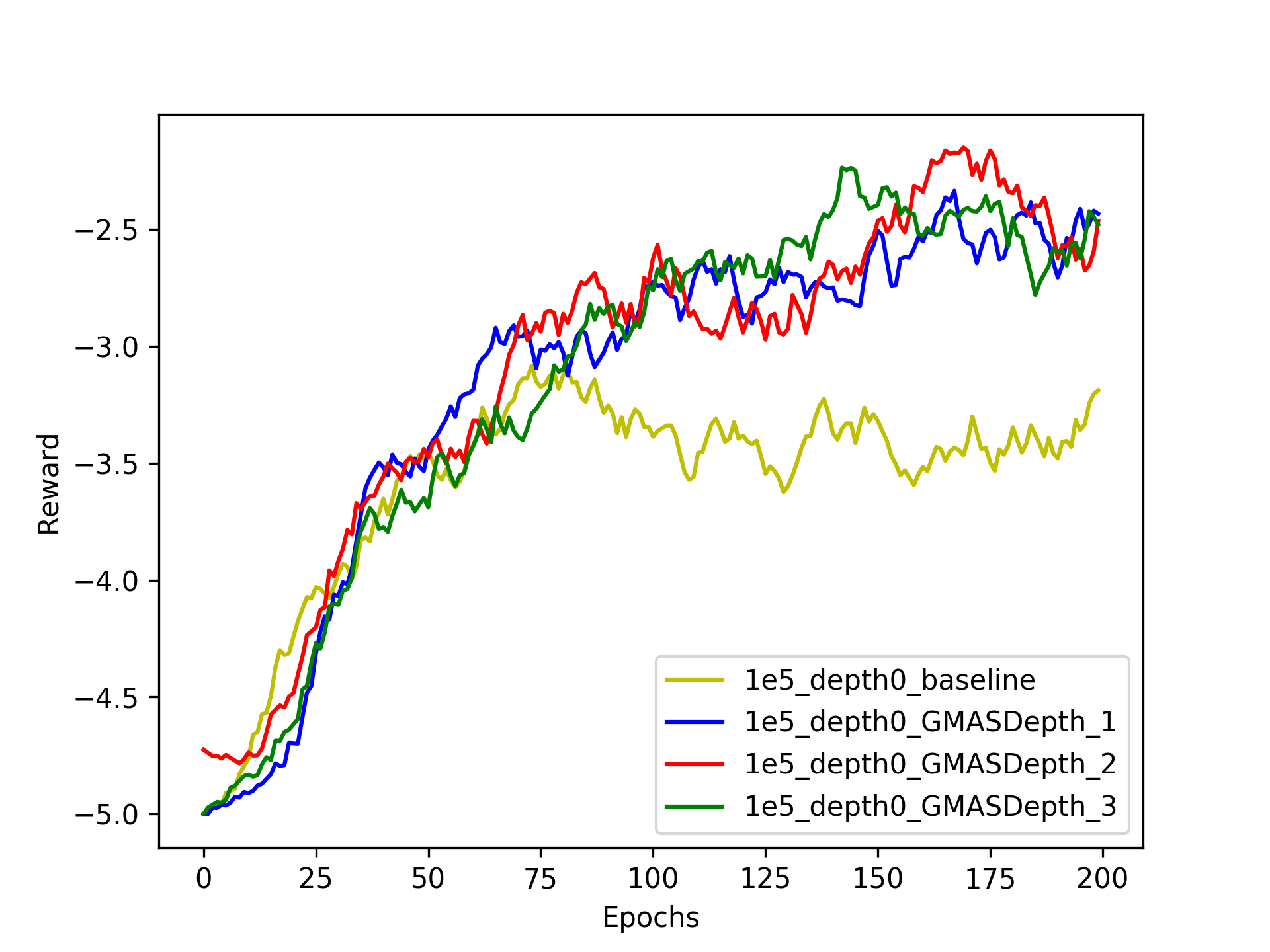}
    \end{subfigure}
    \begin{subfigure}{0.24\textwidth}
    \centering
    \includegraphics[width=\textwidth, height=1.2in]{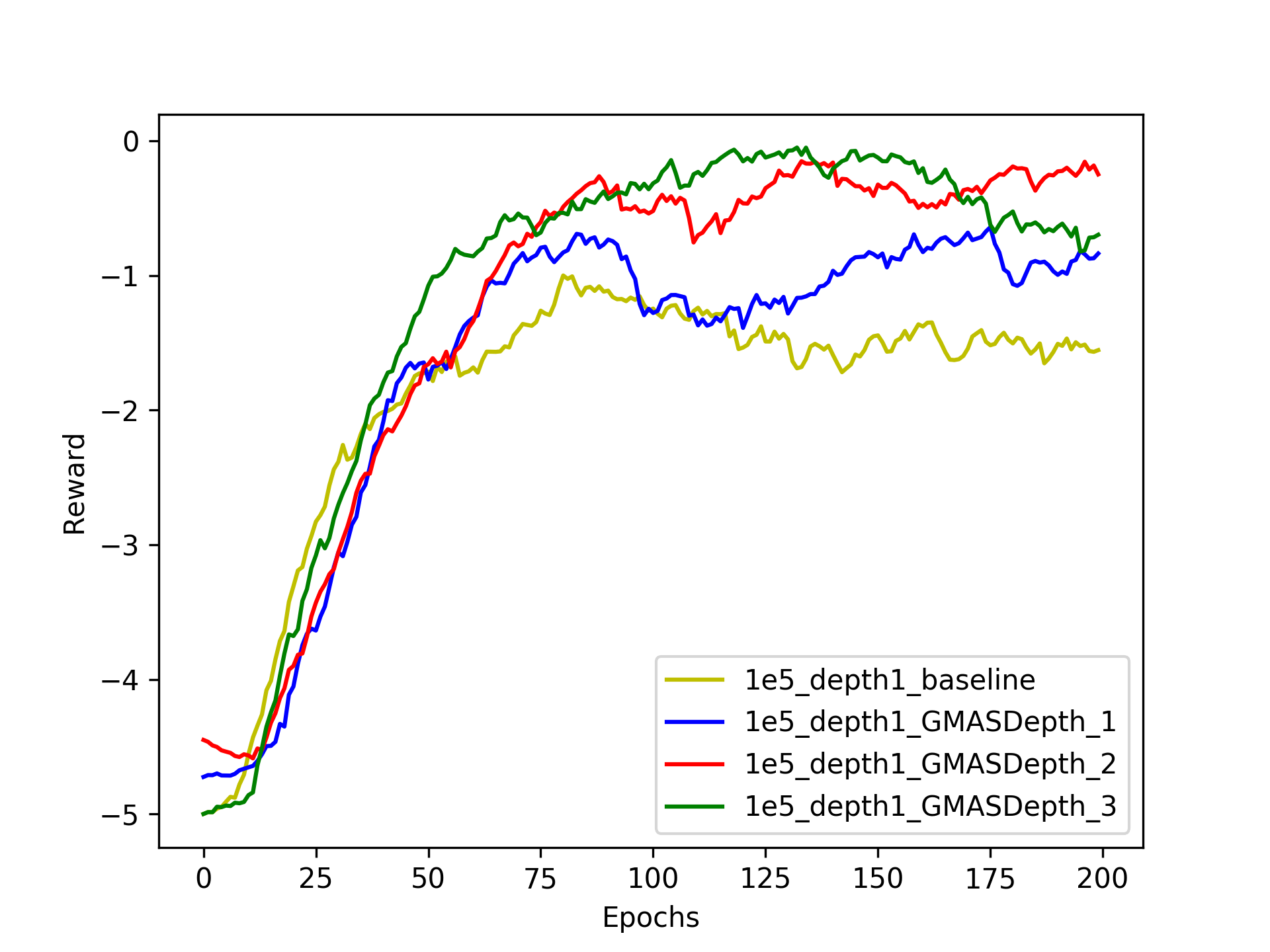}
    \end{subfigure}
    \begin{subfigure}{0.24\textwidth}
    \centering
    \includegraphics[width=\textwidth, height=1.2in]{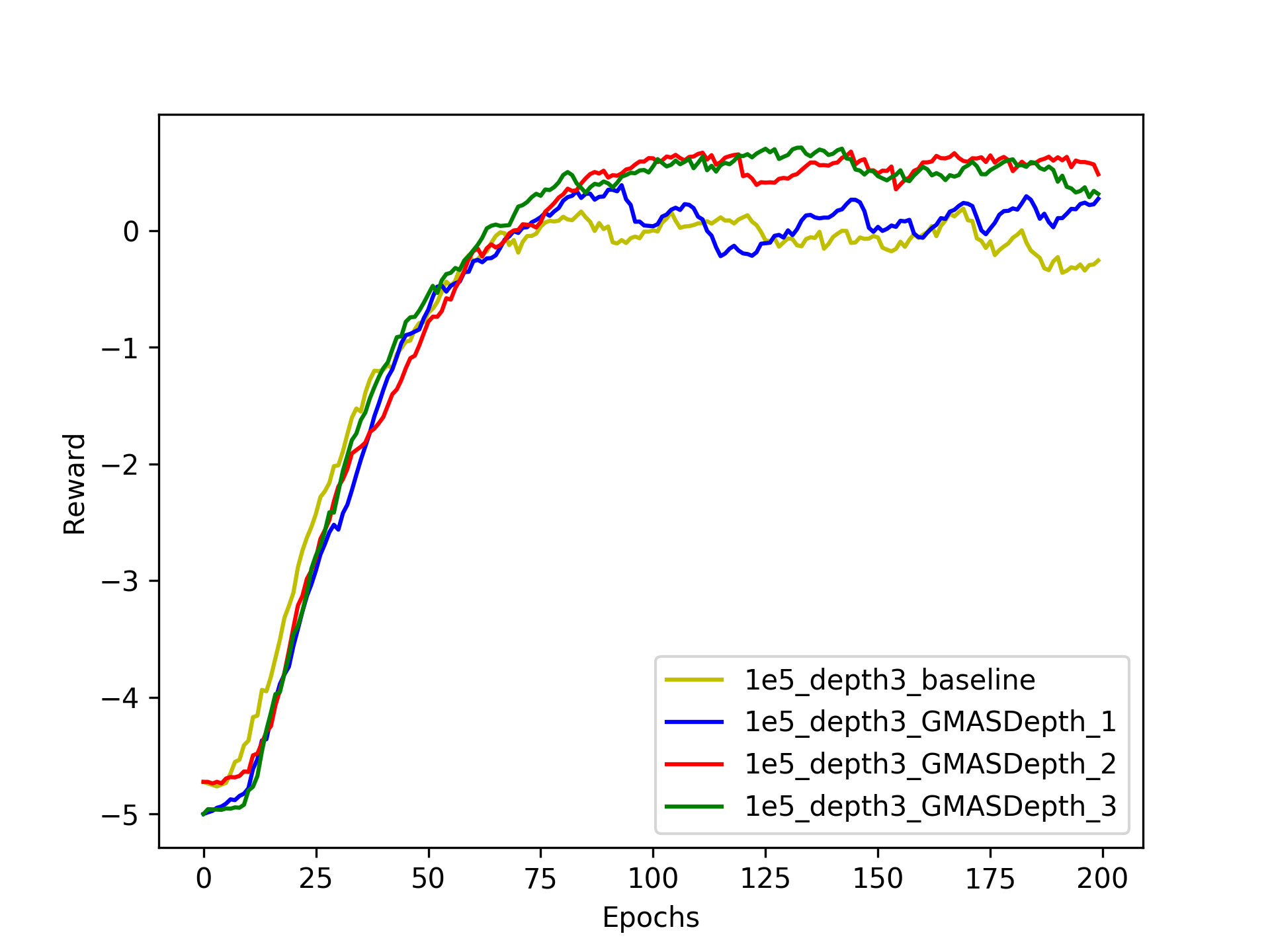}
    \end{subfigure}
    \begin{subfigure}{0.24\textwidth}
    \centering
    \includegraphics[width=\textwidth, height=1.2in]{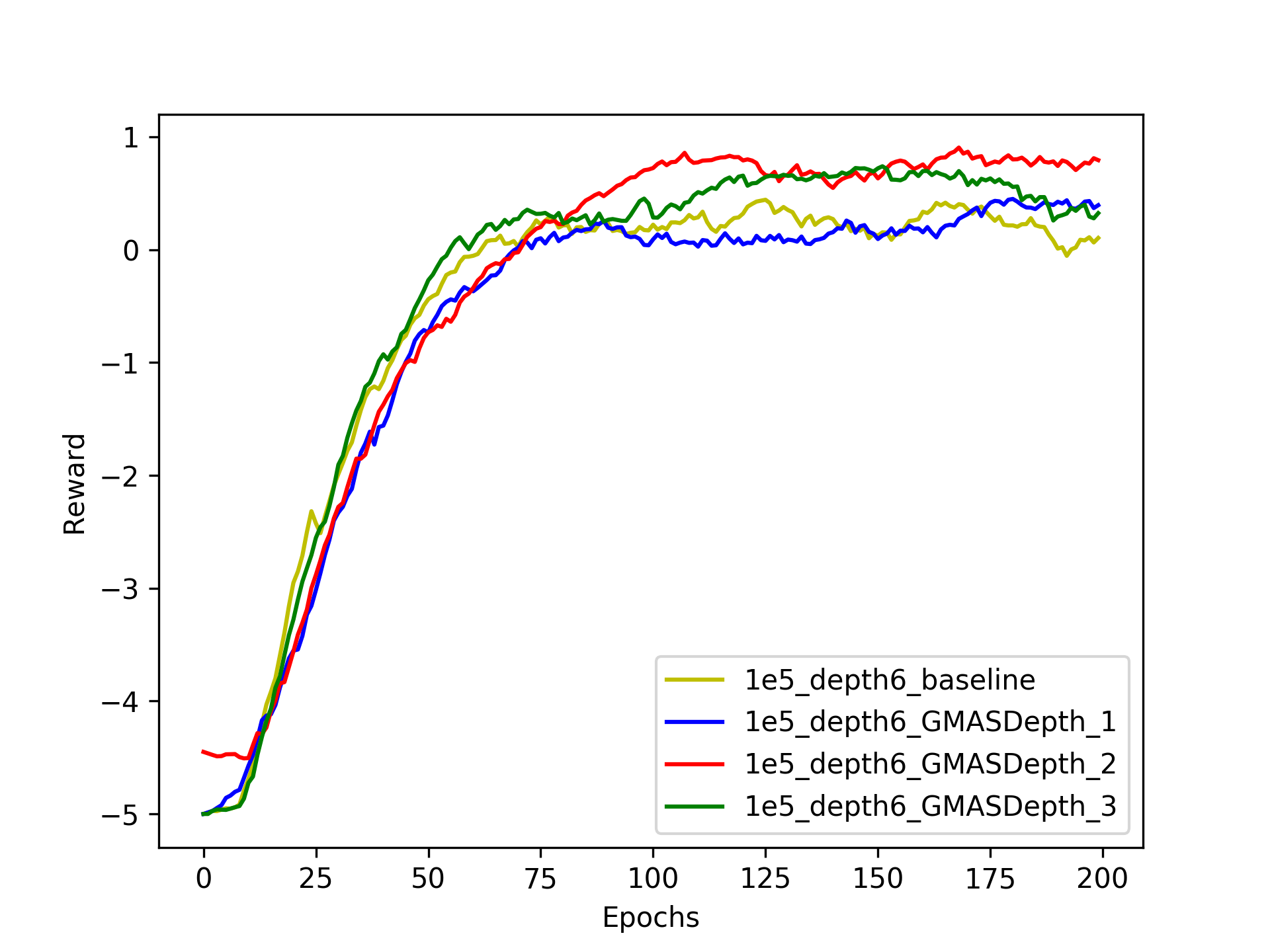}
    \end{subfigure}
   \hspace{1em}% <-- optional space between the subfigures
    \begin{subfigure}{0.24\textwidth}
    \centering
    \includegraphics[width=\textwidth, height=1.2in]{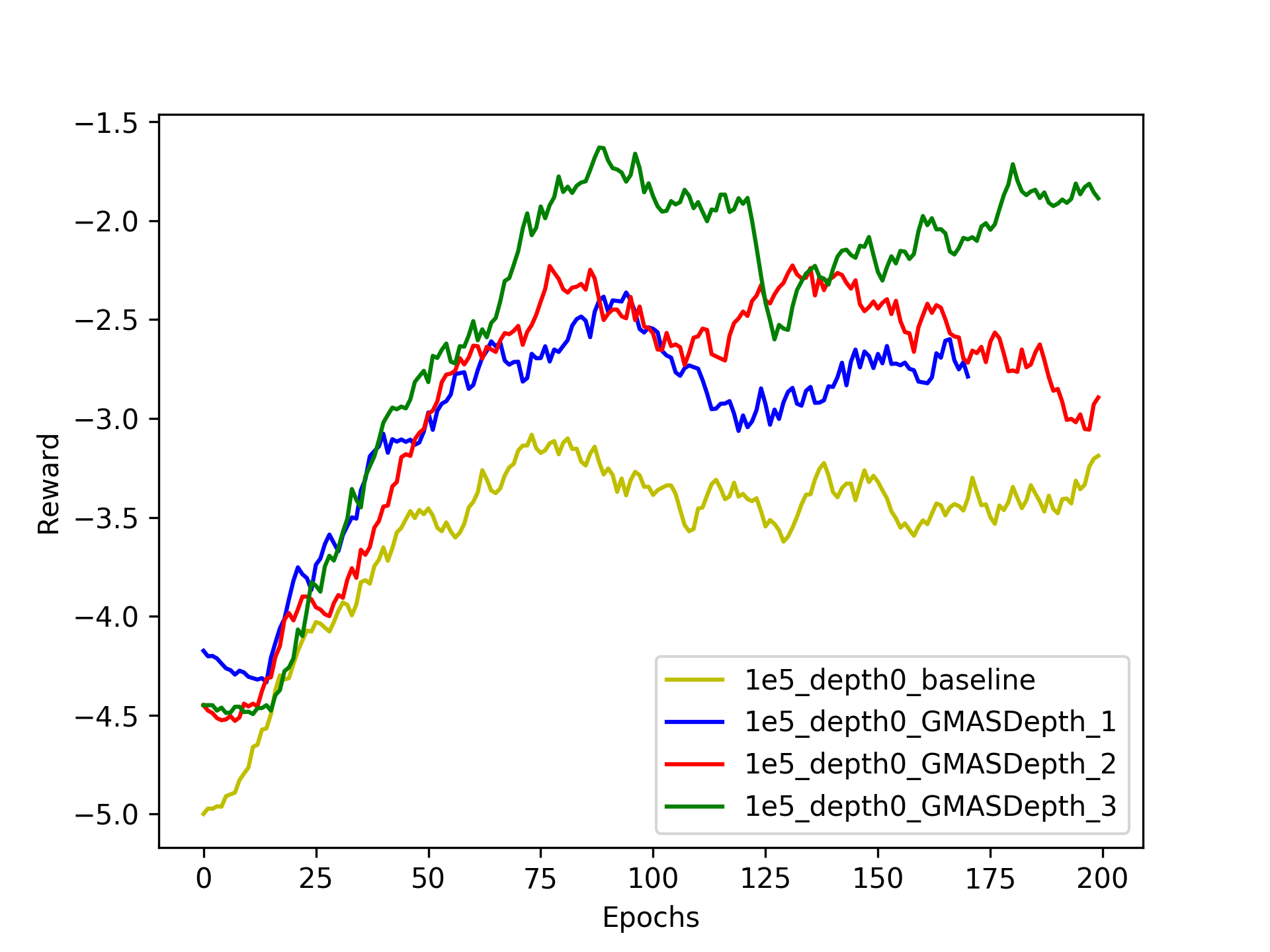}
    \end{subfigure}
    \begin{subfigure}{0.24\textwidth}
    \centering
    \includegraphics[width=\textwidth, height=1.2in]{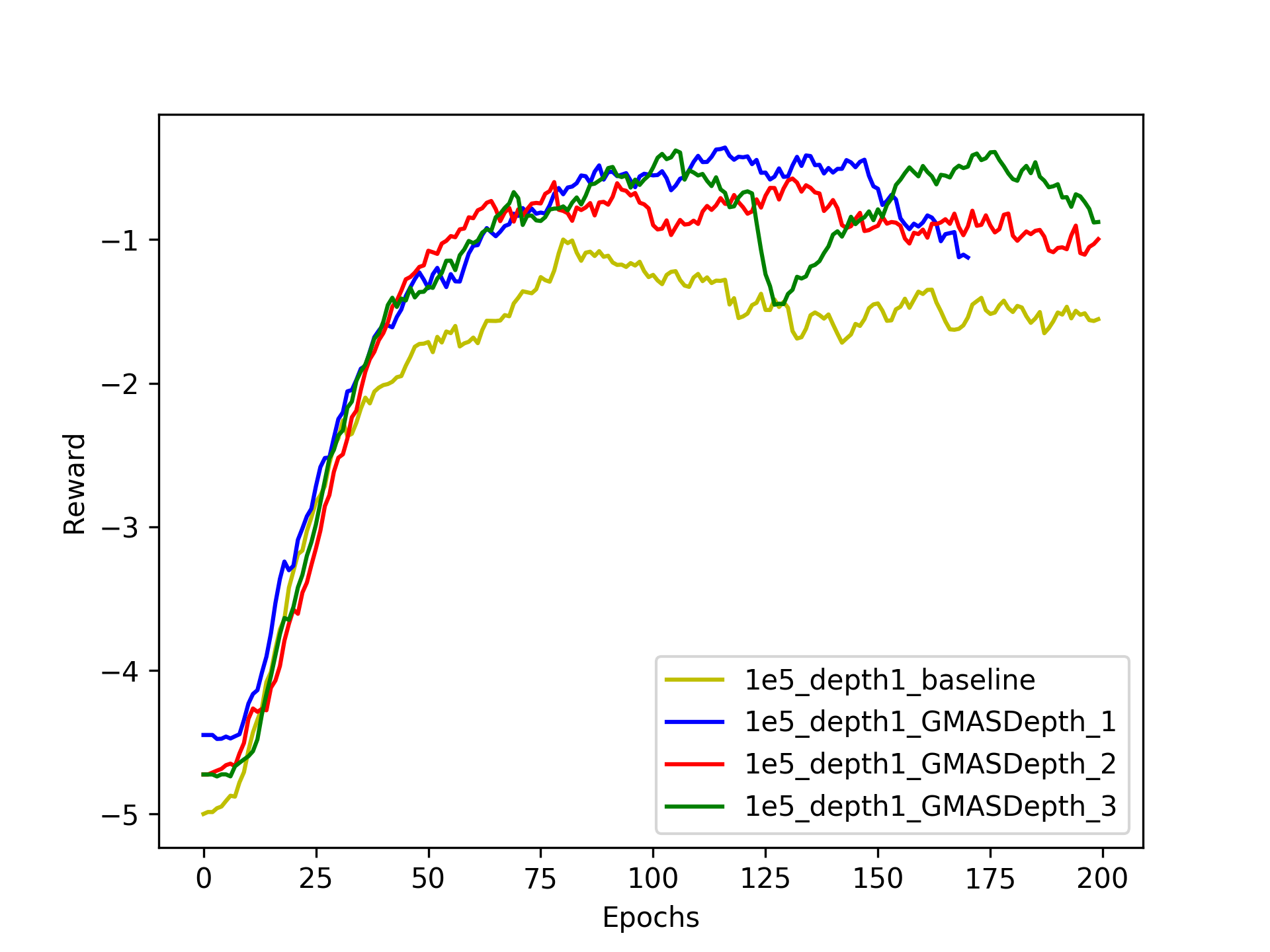}
    \end{subfigure}
    \begin{subfigure}{0.24\textwidth}
    \centering
    \includegraphics[width=\textwidth, height=1.2in]{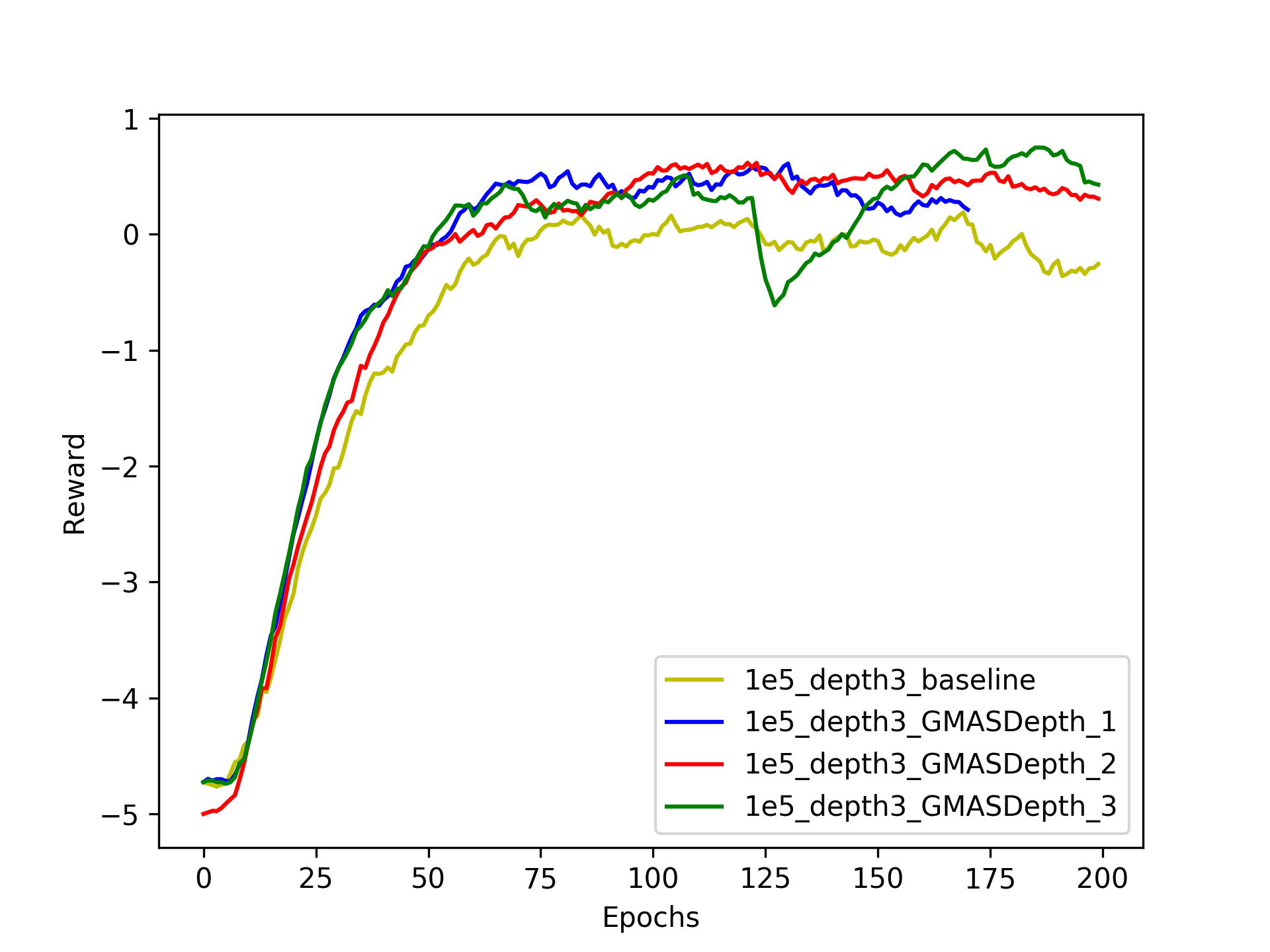}
    \end{subfigure}
    \begin{subfigure}{0.24\textwidth}
    \centering
    \includegraphics[width=\textwidth, height=1.2in]{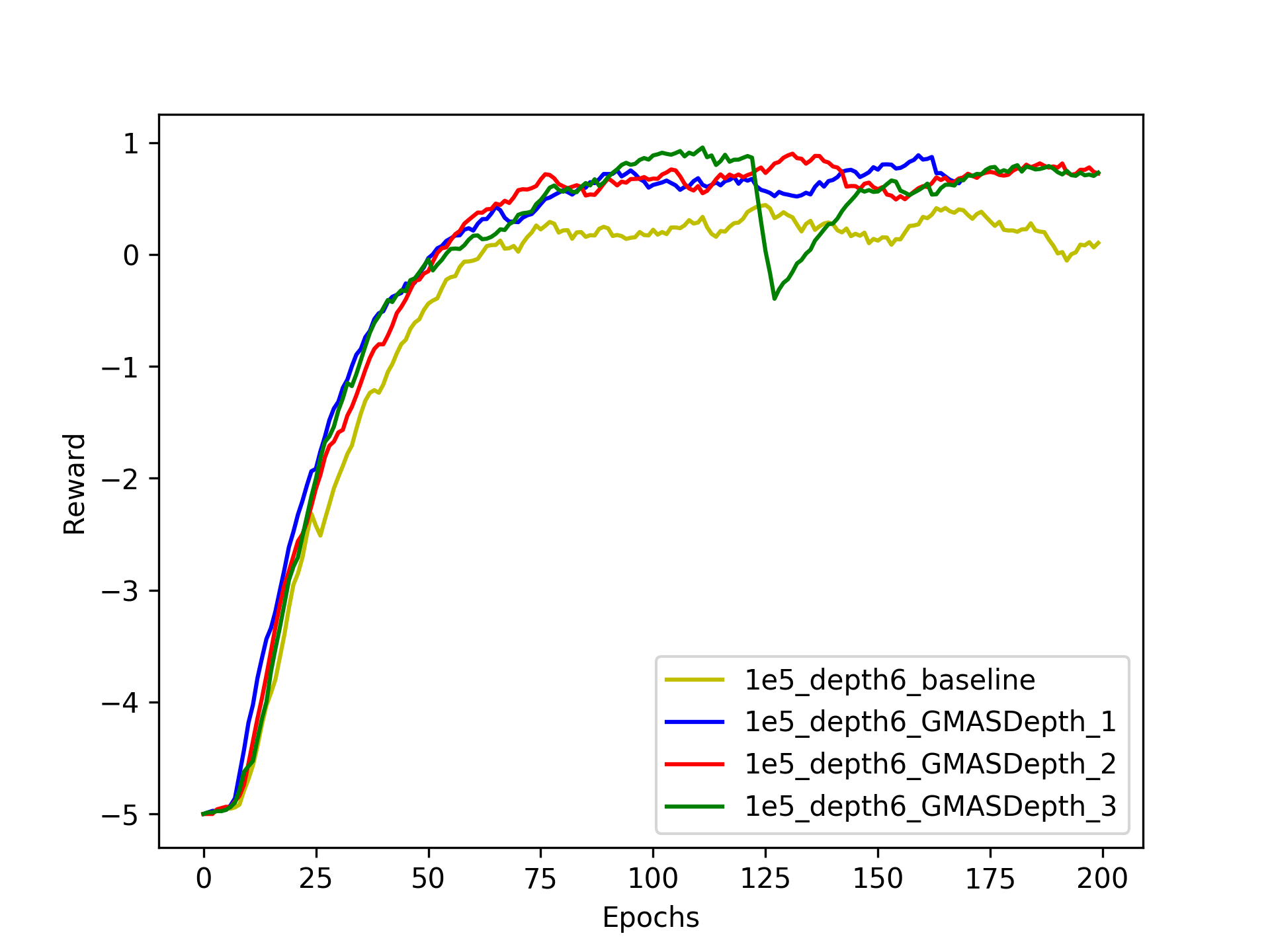}
    \end{subfigure}

   % \centering
    % \subfigure{\includegraphics[width=0.24\textwidth, 
    % height=1.2in]{plot_1e5_depth0_cosine.png}} 
    % \subfigure{\includegraphics[width=0.24\textwidth, height=1.2in]{plot_1e5_depth1_cosine.png}}
    % \subfigure{\includegraphics[width=0.24\textwidth, height=1.2in]{plot_1e5_depth3_cosine.png}} 
    % \subfigure{\includegraphics[width=0.24\textwidth, height=1.2in]{plot_1e5_depth6_cosine.png}}
    % \subfigure{\includegraphics[width=0.24\textwidth, height=1.2in]{plot_1e5_depth0_l2.png}}
    % \subfigure{\includegraphics[width=0.24\textwidth, height=1.2in]{plot_1e5_depth1_l2.png}}
    % \subfigure{\includegraphics[width=0.24\textwidth, height=1.2in]{plot_1e5_depth3_l2.png}} 
    % \subfigure{\includegraphics[width=0.24\textwidth, height=1.2in]{plot_1e5_depth6_l2.png}}
    \caption{Top row: Off-policy data of size $1e5$ with GMAS using cosine distance. Bottom row: Off-policy data of size $1e5$ with GMAS using L2 distance, Evaluation Planning depth varies as [0,1,3,5] from left to right in both rows.}
    \label{fig:foobar}
\end{figure*}

Figure \ref{fig:foobar} demonstrates the results obtained by CRAR (labeled baseline in the plots) and GMAS in the limited off-policy training data setting of $1e5$ samples obtained from the environment. This setting corresponds to the case where CRAR does not solve the environment in the original work. We demonstrate that the additional training signal used by GMAS significantly improves the sample efficiency even in the case of depth 0 (which is equivalent to using only the model free component with no planning). The performance is also shown to improve with planning depth used in GMAS with depth 3 performing better than lower depths in most cases. Note that this is different from the depth used at test time by CRAR and is used to estimate the gradient of the model based learner at training time. It is also worth noting that the performance gains offered by depth are not monotonic and in fact can lead to sub-optimal performance in some cases due to the compounding error problem highlighted in section \ref{compound} and incorporating the mitigation techniques are left as future work. This paper demonstrates the efficacy of the approach in the limited data setting both with respect to sample efficiency and generalization on one set of task distributions and extension of the technique to harder environments and integration with other techniques is an ongoing effort.

% \section{Future work}
\bibliography{iclr2020_conference}
\bibliographystyle{iclr2020_conference}

\appendix
\section{Appendix}
\subsection{Implementation details}
\begin{algorithm}
\label{gradalgo}
\caption{Gradient Estimate of Model-Based Learner}
\begin{algorithmic}[1]

\Procedure{gradAtDepth}{$state, action, depth$}       %\Comment{This is a test}
    % \State System Initialization
    % \State Read the value 
    \If{$depth == 0$}
        \State \Return \textit{gradQ}(state,action)
    \Else
    \State nextState = \textit{predictNextState}(state,action)
    \State bestAction = \textit{getBestActionFromPlanning}(depth)
    \State  
    \textbf{return} \textit{gradReward}(state,action) + \textit{gradGamma}(state,action)*maxQVal +
    \State \hfill gamma*\textit{gradAtDepth}(nextState,bestAction,depth-1)*\textit{gradTransition}(state,action)
    \EndIf
    
    % \If{$condition = True$}
    %     \State Do this
    %     \If{$Condition \geq 1$}
    %     \State Do that
    %     \ElsIf{$Condition \neq 5$}
    %     \State Do another
    %     \State Do that as well
    %     \Else
    %     \State Do otherwise
    %     \EndIf
    % \EndIf

    % \While{$something \not= 0$}  \Comment{put some comments here}
    %     \State $var1 \leftarrow var2$  \Comment{another comment}
    %     \State $var3 \leftarrow var4$
    % \EndWhile  \label{roy's loop}
\EndProcedure
\end{algorithmic}
\end{algorithm}
As discussed in the Section \ref{exp}, we consider tasks sampled from a distribution of maze environments to demonstrate the capability of GMAS. Here, the environment provides a two dimensional state representation of shape $48\times 48$, where each pixel represents a gray-scale value. We use a similar base architecture for CRAR as described in \cite{franccois2019combined} with some notable modifications and additions mentioned below:
\begin{enumerate}
    \item We re-implemented the architecture in PyTorch (\cite{paszke2019pytorch}) for ease of implementation of GMAS and found that a base-learning rate of $1e^{-3}$ improved performance even for the base CRAR agent (original work used $lr=5e^{-4}$).
    \item The scaling factor $\alpha$ from Equation \ref{der} was set to $0.05$ in case of cosine distance loss and $1.0$ in the case of L2 loss. We experimented with $\alpha \in [0.01,0.05,0.1,0.5,1.0]$ for cosine distance and $\alpha \in [0.1, 0.5, 1.0, 5.0]$ for L2 distance and found $0.05$ and $1.0$ respectively to be the optimal values for $\alpha$.
    \item We re-use the same frozen Q-function (frozen for 1000 iterations) for calculation of the gradient estimate of model-based learner.
\end{enumerate} 

\end{document}